\documentclass[conference]{IEEEtran}   

\usepackage[T1]{fontenc}          
\usepackage{newtxtext,newtxmath}  
\usepackage[utf8]{inputenc}       
\usepackage{booktabs}
\usepackage{tabularx}   
\usepackage{graphicx}             
\usepackage{amsmath}              
\usepackage{url}
\usepackage[hidelinks]{hyperref}  

\let\UnderscoreOrig\_
\renewcommand{\_}{\UnderscoreOrig\allowbreak}

\newcommand{\flowspec}{\textsc{FlowSpec}}

\title{Following the Flow the Law Requires:\\
       Training Voice Agents to Transfer to Unseen Regulated Procedures in Debt Collection}

\author{\IEEEauthorblockN{Petmongkon Pornpichitsuwan, Krittin Pachtrachai, Touchapon Kraisingkorn}
\IEEEauthorblockA{Amity Research and Application Center (ARAC)\\
Amity AI Holdings Co.,Ltd., Bangkok, Thailand\\
\{petmongkon, krittin, touchapon\}@amity.co}}

\begin{document}
\bstctlcite{IEEEexample:BSTcontrol}   
\maketitle

\begin{abstract}
Voice agents built on large language models (LLMs) now handle customer calls, but in regulated applications completing the task is not enough: the agent must follow a conversational flow set by law. In debt collection, it must confirm it is speaking to the debtor before mentioning the debt, identify itself and the creditor, state the amount, and avoid prohibited language. Every creditor turns these rules into its own procedure and revises it. We ask whether a small, self-hosted policy can be trained once and then follow regulated procedures it has never seen. One machine-readable specification per deployment, \flowspec{}, generates the policy instruction, gates the tools, supplies the supervision and defines the tests; the agent speaks only approved catalogue sentences. We fine-tune a 9-billion-parameter model on 2,506 conversations from 297 randomized procedures and evaluate a Thai outbound-call agent on four unseen deployment procedures (three debt collection, one non-debt), one from a wholly withheld lineage. \textbf{Documented} checks, fixed by the procedure, rise from 67--71\% to 93--100\%, while \textbf{grounded} checks, fixed by the conversation, already pass at 97--100\% before training; scenario pass rate rises from 15/43 to 40/43. An audit of the shared specification finds 11 defects, 3 of which its own metric did not expose.

\end{abstract}


\section{Introduction}
\label{sec:introduction}

Voice agents built on large language models (LLMs) are moving from demonstrations into contact centres. They hold multi-turn conversations, call backend tools and act on a customer's behalf \cite{yao2023react,barres2025tau2}, and cascaded pipelines of speech recognition, an LLM and speech synthesis now run in real time on enterprise hardware \cite{qiu2026voiceagents}. Research on such agents usually measures whether the user's task gets done. Many customer-facing applications also constrain \emph{how} it gets done. In banking, insurance, healthcare and debt collection, what the agent may say, what it must say, and the order in which it says it are fixed by regulation. For these applications, a capable agent is not enough; the agent must follow a prescribed conversational flow.

Debt collection shows this clearly. Under Thailand's Debt Collection Act \cite{thaidca2015}, a collector who reaches someone other than the debtor may ask only where the debtor can be contacted and may not disclose the debt. When speaking to the debtor, the collector must give its own name, the creditor's name and the amount owed. Calls are allowed only within fixed hours, and threats, profane language and disclosure of the debt to unrelated people are prohibited. United States (U.S.) law imposes similar duties. The collector must disclose in its first communication that it is attempting to collect a debt \cite{fdcpa}, and placing more than seven calls within seven consecutive days is presumed to be harassment \cite{regf2020}. Rules written for artificial intelligence (AI) voices now add to this. AI-generated voices count as ``artificial'' voices under U.S. telephone-consumer law \cite{fcc2024ai}, and the European Union (EU) AI Act requires that people be told when they are talking to an AI system \cite{euaiact2024}. Within a call, these rules define a \emph{flow}: identity must be verified before the debt is disclosed, the mandated disclosures must be made, and prohibited statements must never be made. Creditors add their own steps on top, such as limits on repeated payment requests and the disposition under which each call is recorded. An agent that says the right things in the wrong order is still non-compliant.

Neither of the two established ways of building such agents is satisfactory. A free-form LLM agent cannot guarantee the flow. Language models produce fluent but unsupported content \cite{ji2023hallucination}, a tribunal has held a company liable for what its chatbot told a customer \cite{moffatt2024}, and tool-using agents given a written policy at inference time still follow it unreliably \cite{yao2024taubench,barres2025tau2,xiao2024flowbench,sopbench}. Traditional flow-based chatbots and voicebots guarantee the flow by construction, and collection systems are commonly built this way \cite{luo2025debtretrieval,qian2022outbound}. But their hand-authored intent classifiers and state machines break on the variety of real calls, and every new client means authoring and maintaining another flow (Section~\ref{sec:related-flow}). We keep the guarantee of the flow-based approach without its brittleness. In our system the agent never generates spoken language: a learned policy selects entries from a catalogue that the client has approved and calls declared tools, so compliance review covers a finite set of sentences, while the policy handles the conversation.

This leaves the problem of following each client's flow. The legal requirements are shared, but each creditor turns them into its own procedure, with its own disclosure order, tool names, payment-request limits, disposition codes and definition of a successful call, and it revises that procedure over time. A policy trained for one procedure must be retrained and revalidated when the procedure changes, and it does not carry over to a new client. Prompting a hosted model with the procedure does not suit this setting either. Collection calls carry personal and financial data that clients keep on their own premises, and Thailand's Personal Data Protection Act restricts transfers of such data abroad \cite{thaipdpa2019}. The agent also runs in a cascaded speech pipeline with a budget of about one second per spoken turn, and one turn can involve several policy calls. The client hardware provides a single 40\,GB accelerator. The cascaded design remains the practical choice for self-hosted real-time voice agents, because end-to-end speech models are not yet self-hostable at usable latency \cite{qiu2026voiceagents}, and it keeps the action-level control that compliance requires. The policy must therefore be small, run on the client's own hardware, and follow each client's procedure without per-client training.

We therefore ask: \emph{can a small, self-hosted dialogue policy be trained once and then follow regulated procedures it has never seen, while speaking only from each client's approved catalogue?}

\subsection{This Paper}

We put the regulated procedure inside the training and evaluation loop. Each deployment is described by one machine-readable specification, \flowspec{}. The same document generates the policy instruction, declares and gates the tools, defines the reference walk used for supervision, and specifies the evaluation checks. We fine-tune one 9-billion-parameter (9B) policy with low-rank adaptation (LoRA) \cite{hu2021lora} on 2,506 conversations from 297 generated procedures derived from 7 base lineages. Identifiers, tool names and flow structure are randomized per example, and one complete lineage is held out.

To see what the policy learns, we sort evaluation checks by what determines the expected behaviour. A check is \textbf{documented} when the procedure fixes its expected value (for example, the required tools and the closing disposition) and \textbf{grounded} when the conversation fixes it (for example, whether an argument matches what the caller said). Before training, the policy passes 97--100\% of grounded checks but only 67--71\% of documented ones. Training raises documented performance to 93--100\%, including on the withheld lineage. In this evaluation, following a regulated flow behaves as an ability separate from conversational competence, and it is where training helps most. A larger untrained model with 27 billion parameters (27B) passes 37/43 scenarios but does not fit that hardware at 16-bit precision; training brings the 9B policy that does fit from 15/43 to 40/43.

Our contributions are:
\begin{itemize}
  \item A formulation of regulated voice-agent behaviour as \emph{procedure following}, in which one machine-readable specification drives the instruction, the tool contract, the supervision and the tests, and speech is restricted to an approved catalogue.
  \item Evidence from a Thai outbound debt-collection agent that one 9B policy, trained on randomized procedures, follows procedures from a lineage it never saw (documented checks 67\%$\rightarrow$93\%; scenarios 8/21$\rightarrow$18/21).
  \item A split of procedure following into documented and grounded components, showing that training gains are concentrated in the documented component.
  \item An audit of the shared specification that finds 11 defects, 3 of which the derived metric cannot detect, and so identifies a blind spot of specification-conditioned evaluation.
\end{itemize}

Section~\ref{sec:related} reviews related work, Section~\ref{sec:method} describes the method, Section~\ref{sec:experiments} reports the experiments, Section~\ref{sec:discussion} discusses what the results establish and their limitations, and Section~\ref{sec:conclusion} concludes.

\section{Related Work}
\label{sec:related}

\subsection{Why Traditional Flow-Based Chatbots and Voicebots Fall Short}
\label{sec:related-flow}

\textbf{Hand-authored flows.} Spoken dialogue systems have long been built on explicit dialogue control: finite-state designs in which the system leads the caller through a fixed network of prompts, and frame-based designs in which the system fills a set of slots \cite{mctear2002}. Industrial chatbot and voicebot frameworks follow the same pattern. A natural language understanding (NLU) component maps each utterance to one of a fixed set of intents, and a hand-authored flow decides the next step; this intent-based design remains the predominant one in industry \cite{bocklisch2024calm}. Surveys of task-oriented dialogue describe this modular pipeline and its reliance on hand-crafted components and domain-specific annotation \cite{zhang2020todsurvey,gao2018neural}. For compliance, the design has a real strength: every path is written in advance and can be reviewed.

\textbf{Brittleness to real conversation.} The same design fails when calls leave the authored paths. An intent classifier must be trained with annotated examples for every intent in every new domain \cite{casanueva2020intent}, and it assumes that each utterance belongs to a supported intent, although real users regularly say things outside that set \cite{larson2019oos}. Speech makes this worse. Automatic speech recognition (ASR) errors propagate into spoken language understanding (SLU) \cite{simonnet2017asr}, and statistical dialogue management was motivated precisely by the cost of hand-crafting dialogue managers and their lack of robustness to recognition errors \cite{williams2007pomdp,young2013pomdp}. In deployed voice systems, breakdowns come from both speech-to-text mistranscriptions and user flows the designers did not anticipate \cite{miah2024breakdown}, and modular interactive voice response (IVR) systems suffer from accumulated errors across components and a high maintenance overhead \cite{zhang2026duivrs}.

\textbf{Cost of one flow per client.} Each new deployment requires authoring, annotating and testing another flow, and each procedure revision requires editing it. For debt collection, where every creditor has its own procedure, this cost grows with the number of clients. Hybrid systems reduce it by keeping deterministic business logic while letting a learned model handle the conversation \cite{williams2017hcn,bocklisch2024calm}. LLMs, in turn, handle open conversation well but track explicit dialogue state less reliably than specialised models, although they can drive a dialogue to success when given the correct state \cite{hudecek2023llmtod}. Our approach sits between the two. The flow stays explicit, as data in \flowspec{}, which keeps the reviewability of a flow-based system, while a learned policy handles the conversation. The documented--grounded split in Section~\ref{sec:checkclasses} measures exactly these two sides.

\subsection{Learned and LLM-Based Approaches}
\label{sec:related-learned}

\textbf{Flow- and schema-grounded dialogue.} Task-oriented dialogue research has long conditioned agents on explicit task structure. Schema-guided datasets and models \cite{rastogi2020sgd,mehri2021sam} give the agent a description of an unseen task, and the Action-Based Conversations Dataset (ABCD) \cite{chen2021abcd} requires agents to follow multi-step, policy-constrained action sequences in customer service. FloDial \cite{raghu2021flodial} grounds dialogue in troubleshooting flowcharts and tests transfer to flowcharts not seen in training, which is the closest precedent to our transfer question. These systems generate free text and report aggregate task success. That metric does not show whether a failure comes from misreading the procedure or from misreading the conversation, and our setting needs to tell these apart.

\textbf{Procedures given to LLM agents at runtime.} Recent benchmarks evaluate tool-using agents against domain policies and standard operating procedures (SOPs) \cite{yao2024taubench,barres2025tau2,sopbench,xiao2024flowbench}. Agent frameworks supply these procedures as natural language or a procedure language at inference time \cite{shi2025flowagent}. A different approach compiles policy documents into the model's weights through continued pretraining \cite{capcpt}, which makes every procedure change a model update. In all of these, the procedure describes the desired behaviour, but the training signal and the test suite are written separately, so nothing ensures that the three describe the same behaviour.

\textbf{Enforcing compliance.} Guardrail toolkits filter inputs and outputs \cite{rebedea2023nemo}. Policy-compliance work compiles policies into guard code checked before each tool call \cite{zwerdling2025policy}, verifies actions against the dialogue \cite{policyguard}, or checks action traces against temporal-logic specifications \cite{agentltl}. Constrained decoding restricts output to a grammar or schema \cite{willard2023outlines}. We enforce the flow through the action space instead. The policy chooses from an enumeration of approved catalogue identifiers, which follows the response-selection line of dialogue systems \cite{henderson2019responsesel,williams2017hcn}. Unsafe language is therefore excluded by construction, and the remaining question is whether the right entry is chosen at the right point in the flow.

\textbf{Training agents from an oracle.} Tool-using LLM agents \cite{yao2023react} are commonly trained on verified or simulated trajectories \cite{apigenmt}. Related work examines error compounding in imitation \cite{dagger} and how supervised fine-tuning and reinforcement learning differ in generalization \cite{chu2025sftrl}. These studies train or evaluate agents against a given specification. Our question is different: when the specification itself supplies the supervision, does the policy learn to follow procedures that are absent from training? Answering this requires holding out whole procedure lineages, not individual examples.

\textbf{Dialogue systems for debt collection.} Prior work on collection dialogue studies negotiation strategy and recovery \cite{wang2025dcn,yang2026debtbench}, strategy selection for outbound robots \cite{qian2022outbound}, retrieval of human-written collection scripts \cite{luo2025debtretrieval}, and conversation understanding in collection call centres \cite{nguyen2026creditcgpt}. None of this work tests whether an agent follows a legally mandated flow, or whether that ability transfers across creditors' procedures.

\textbf{Validity of the evaluation.} If one procedure generates both the supervision and the test criteria, a defect in that procedure affects both, and the policy can score perfectly while violating the intended behaviour. Failure attribution for agents \cite{whoandwhen} and evidence that standard metrics can hide policy failures \cite{retrievalmislead} motivate auditing the specification and the tests, not only the policy.

\section{Proposed Method}
\label{sec:method}

The problem identified in Section~\ref{sec:introduction} is that the procedure typically sits outside the loop that trains and evaluates the policy. Our method places the procedure inside this loop. For each deployment, we define a single machine-readable document that contains the deployment-specific rules. The same document is consumed in four ways: it generates the policy instruction, defines the environment and its tool contract, generates the supervision targets, and defines the evaluation checks. The policy is then trained across a generated population of such procedures, with complete procedure lineages held out from training. This makes following an unseen procedure an explicit part of the training and evaluation design.

\subsection{\flowspec{}: One Document, Four Consumers}
\label{sec:flowspec}

\begin{figure*}[t]
  \centering
  \includegraphics[width=0.82\textwidth]{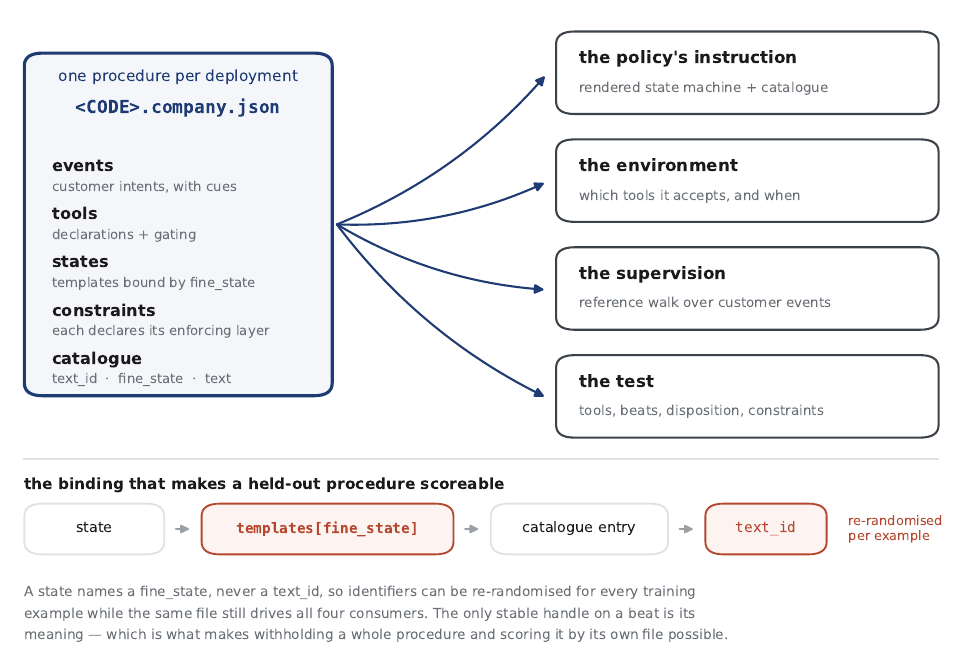}
  \caption{One procedure and its four consumers. A state names a semantic label rather than a catalogue identifier, so the identifier space can be permuted for every training example while the same document drives the instruction, the environment, the supervision and the test.}
  \label{fig:onedoc}
\end{figure*}

The central design principle is that the four consumers are not independently maintained artefacts. They are four representations derived from the same machine-readable procedure (Fig.~\ref{fig:onedoc}). This provides a single object that can be held out during evaluation and ensures that the instruction, environment, supervision, and evaluation criteria are derived from the same source.

\subsubsection{Procedure Representation}
\label{sec:representation}

Each deployment is represented by one JavaScript Object Notation (JSON) document loaded through a common interface. A representative procedure contains 15 states, 7 tools, 15 customer events, 35 constraints, 16 frequently asked question (FAQ) routes, and a 57-entry utterance catalogue. Appendix~\ref{app:spec} gives the schema, validation rules and an example.

Three components carry most of the behavioural information: states, tools, and constraints.

States define the dialogue flow. Each state specifies its phase, the utterances allowed in that phase, the tools that must run on entry, and a transition for each customer event. A terminal state may also specify an outcome and the disposition under which the call should be recorded.

States do not directly reference catalogue entries by their identifiers. Instead, they reference a semantic \texttt{fine\_state} label that is resolved against the catalogue at rendering time. This indirection is enforced by the validator and allows catalogue identifiers to be randomized without changing the underlying procedure (Section~\ref{sec:randomization}).

Tools are explicitly declared in the procedure rather than assumed by deployment-specific code. Each tool specifies its name, argument schema, and gating conditions. For example, an account-read tool may require the caller to confirm their name before it can be called, require the tool to run before a state that discloses the account balance, and limit its use to once per call. The environment reads these declarations directly, so no company-specific tool branches are required.

Constraints represent rules that cannot be expressed directly by the flow graph. Each constraint specifies the layers responsible for enforcing it, such as the prompt, reward, backend, or session. For example, a maximum number of payment requests can be represented as a typed constraint with a transition that is taken when the limit is exceeded.

In the representative deployment, only 8 of 35 constraints are typed and machine-checkable; the remaining 27 are prose constraints presented to the model without automatic verification. Thus, the enforcement-layer field identifies where a constraint is intended to be enforced but does not make a prose constraint automatically enforceable. The evaluation in Section~\ref{sec:experiments} therefore covers only typed constraints.

The validator is part of this design. Because the same document feeds multiple consumers, inconsistencies that could otherwise appear silently in one component must instead be rejected during validation. Appendix~\ref{app:spec} lists these validation rules.

\subsubsection{Procedure $\rightarrow$ Instruction}

A renderer converts the procedure into a state-machine instruction driven by customer events. FAQ routes can interrupt any non-terminal state and resume the previous flow afterward. Constraints marked for prompt enforcement are included as explicit rules, and the utterance catalogue defines the vocabulary from which the policy can select responses.

Training and serving may package the catalogue differently, but both use the same rendering core. Its contract is that the relevant procedure rendering is byte-identical between training and serving. This prevents the policy from being trained against one representation of a procedure and evaluated against another.

\subsubsection{Procedure $\rightarrow$ Environment}
\label{sec:env}

The environment is deployment-agnostic: it does not contain company-specific behavioural branches. Instead, it reads the declarations in the procedure. Tool schemas are generated from the procedure, including argument names, types, enumerations, formats, and required fields. Session-level requirements, such as identity verification, consent before an action, and date-time requirements before a commitment, are likewise derived from the procedure structure.

The policy uses a structured action space throughout the call:

\begin{equation*}
a_t = (\text{tool}, \text{arguments}).
\end{equation*}

This includes spoken responses. The policy does not generate free-form text. Speaking is represented as a call to a reply tool whose argument is a list of catalogue identifiers. The identifiers are declared as an enumeration containing exactly the catalogue entries available in that conversation, in the spirit of response selection over a fixed candidate set \cite{henderson2019responsesel}. The environment resolves the selected identifier, fills any required slots using its own records, and renders the corresponding approved utterance (Appendix~\ref{app:spec}).

Consequently, an approved catalogue entry is the only text that can be spoken by construction. Compliance review therefore applies to a finite catalogue of approved sentences rather than to arbitrary model-generated language. This is what makes the legal requirements of Section~\ref{sec:introduction} checkable: a mandated disclosure is a catalogue entry that must be selected, and a prohibited statement is one that the catalogue does not contain.

This design also separates deployment-specific behaviour from policy parameters. The policy selects canonical actions, while the environment maps those actions to the deployment-specific tool names, limits, disclosure order, and closing rules defined by the procedure. Therefore, a procedure with different deployment rules can be executed without adding company-specific branches to the policy environment.

\subsubsection{Procedure $\rightarrow$ Supervision}

A procedure walker traverses the same state machine using a sequence of customer events. For each event, it returns the tools that are expected in order, the utterances allowed at that point, whether an FAQ response is pending, and whether the resulting state is terminal and, if so, which outcome applies.

The reply obligation is represented as a set rather than a single required sentence. Evaluation therefore checks whether an appropriate catalogue entry is selected rather than requiring exact equality with one predetermined phrasing. This allows multiple valid phrasings and FAQ-resume behaviour.

The customer events determine which branch of the procedure is taken. The walker only determines what behaviour is expected given those events; it does not determine whether the policy correctly identifies the branch.

An important consequence is that the same specification-derived oracle generates both the supervision targets and the reference used by the evaluation metric. This shared dependency is examined in Section~\ref{sec:discussion}.

\subsubsection{Procedure $\rightarrow$ Test}
\label{sec:test}

Each evaluation scenario contains a fixed customer script and an expectation block. A shared checker evaluates the tools required by the procedure, the allowed beats along the resulting walk, the terminal disposition, and the deployment's typed constraints.

The test suite also supports negative tool expectations: a scenario may specify tools that must not be called for that case. Five of the 43 scenarios contain such expectations. None of these expectations fires in either evaluation arm reported in Section~\ref{sec:experiments}, so they do not contribute to the reported difference between the arms.

All other evaluation criteria are derived from the procedure itself. This is what allows held-out procedures to be evaluated without writing a separate deployment-specific answer key or evaluation implementation for each procedure.

\subsection{Generating a Population of Procedures}
\label{sec:population}

Transfer to an unseen procedure is meaningful only if the training set does not contain a closely related version of that procedure. We therefore train on a population of generated procedures rather than on a single procedure and evaluate on individual files alone.

Eight base lineages are used to generate the full population of 340 procedures: four debt-collection deployments and four hypothetical outbound-call scenarios covering appointment reminders, delivery notifications, telephone surveys, and entertainment-related calls. These base lineages are used to construct the synthetic procedure population; no real company appears in the training or evaluation data.

A mutation pass generates synthetic procedures by rewriting the procedure document rather than perturbing its rendered prompt. Nine variation levers modify aspects such as tool names, argument names, identity-verification requirements, payment-request limits, account-read gating, callback branches, FAQ coverage, catalogue phrasing, and optional tools (Table~\ref{tab:i}).

\begin{table*}[!ht]
\centering
\caption{Variation levers across generated procedures. The probabilities are the generator's current values; no sweep was run over them.}
\label{tab:i}
\footnotesize
\begin{tabular}{@{}ll@{}}
\toprule
Lever & How it varies \\
\midrule
Tool names & Drawn from a pool keyed by the tool's role \\
Argument names & Surface names redrawn, mapped back \\
Identity check first & p = 0.4 \\
Payment-request quota & Drawn from {1, 2, 2, 3} \\
Account-read gating & p = 0.35 \\
Callback branch dropped & p = 0.3 \\
FAQ subset & 0--4 optional intents dropped \\
Extra phrasings per beat & max(4, 25--60\%) of beats, always \\
Bonus generic tool & p = 0.35 \\
\bottomrule
\end{tabular}
\end{table*}

The phrasing variation is intentionally applied at the procedure level rather than being fixed for a particular beat. Different procedures therefore contain different numbers of alternative phrasings, preventing the policy from learning a fixed mapping between a beat and a particular number or pattern of responses. The catalogue provided with each example remains the source of the permitted response choices.

\subsection{Generating the Conversations}

The agent side of the conversation generator is deterministic with respect to the procedure: it walks the procedure's state machine rather than using an LLM to generate agent behaviour. Each generated trajectory is therefore consistent with the state machine by construction rather than through post-hoc filtering.

The walker does not automatically guarantee all declared constraints. Conversations that violate a constraint are re-scored and removed before training. This distinction is important because the resulting training data is guaranteed to follow the flow structure, while constraint compliance is separately checked.

The customer side samples event paths with a bias toward transitions that have received less coverage. This encourages the generated population to exercise different branches rather than repeatedly following the happy path. During generation, tool calls are dispatched through a real backend. A call that the production backend would reject therefore becomes a generator failure instead of silently entering the training data. Reject-then-retry sequences are injected with a small probability and recorded in the example metadata.

\subsection{Per-Example Randomization}
\label{sec:randomization}

Two properties are randomized independently for every generated conversation rather than once per procedure: catalogue identifiers are renumbered and their order is shuffled, and the procedure instruction rendered for the example is regenerated from the corresponding procedure.

This randomization is enabled by the structured action space described in Section~\ref{sec:env}. Because speaking requires selecting an identifier from an enumeration, a stable identifier could become a shortcut that allows the policy to select an answer without using the catalogue. Reordering and renumbering the identifiers removes this shortcut. The policy must instead use the catalogue supplied with the current example.

The binding between states and \texttt{fine\_state} labels is therefore essential: states refer to semantic labels rather than catalogue identifiers, allowing the identifier space to change without changing the underlying procedure.

\subsection{Lineage-Level Holdout}

We hold out complete procedure lineages rather than individual procedure files. All procedures derived from a withheld base lineage are excluded from training. Holding out only one mutated file would not provide a strong transfer test because closely related procedures derived from the same base could remain in the training data.

The full generated population contains 340 procedures derived from eight base lineages (Table~\ref{tab:ii}). After withholding one complete lineage, 297 procedures from the remaining seven lineages are used for training.

\begin{table}[h]
\centering
\caption{The generated population is 8 base lineages and 340 mutations of them, not 340 independent procedures. Mean tool-name overlap with a procedure's own base is 0.36 and the minimum is 0; 85 of the 340 differ from their base in state count, 86 in transition count, and 274 in FAQ-route count.}
\label{tab:ii}
\footnotesize
\begin{tabular}{@{}lrrr@{}}
\toprule
Over the 340 generated & Min & Max & Mean \\
\midrule
States & 4 & 16 & 9.4 \\
Transitions & 7 & 50 & 20.0 \\
Declared tools & 1 & 9 & 4.1 \\
Typed constraints & 1 & 12 & 5.4 \\
FAQ routes & 1 & 16 & 7.2 \\
Catalogue entries & 13 & 104 & 44.8 \\
\bottomrule
\end{tabular}
\end{table}

The final evaluation contains no procedure derived from the withheld lineage. Of the four deployment procedures evaluated in Section~\ref{sec:experiments}, three come from lineages represented in training, while the fourth belongs to the withheld lineage.

\subsection{Training Run}

From the 297 training procedures, the generator produces 2,637 conversations, of which 5\% are reserved for validation, leaving 2,506 for training.

Training uses supervised fine-tuning with token-level cross-entropy over the generated trajectories. A LoRA adapter \cite{hu2021lora} is trained while the base model remains frozen. The loss mask covers assistant turns only; the rendered procedure, customer turns, and backend tool results are provided as context rather than targets.

Because the action space consists entirely of structured tool calls, every supervised token belongs to a tool call. For spoken responses, the model is trained to select catalogue identifiers rather than to reproduce the rendered sentence itself. The environment subsequently resolves the selected identifier to the approved utterance.

No preference data or policy-gradient optimization is used. The enforcement layer specified for a constraint determines where that constraint is checked; it does not change the supervised fine-tuning objective.

We use Qwen3.5-9B with a rank-32 LoRA applied to all linear layers. The model is trained for three epochs over 468 steps using one NVIDIA H100 graphics processing unit (GPU). Training takes approximately five hours and produces a 205 MB adapter. The complete configuration and training losses are provided in Appendix~\ref{app:training}.

\subsection{Scope of the Design}

The procedure randomization described above is the mechanism used to encourage transfer across procedures. However, the current study does not include a non-randomized control condition. We therefore report the transfer obtained with the proposed design but do not attribute the observed improvement specifically to randomization without an ablation.

The study establishes whether a policy trained across the generated procedure population can follow procedures from held-out lineages. It does not isolate the contribution of each individual component of the design. These limitations are discussed further in Section~\ref{sec:discussion}.

\section{Experiments and Results}
\label{sec:experiments}

``Following a procedure'' is not a single measured behaviour in our evaluation. We therefore separate checks according to what determines the condition being evaluated. \textbf{Documented checks} test behaviour determined by the procedure, such as which tools a state requires and which disposition closes the call. \textbf{Grounded checks} test behaviour determined by the observed conversation, such as whether a tool argument matches what the caller said. We report the two groups separately for deployments whose related lineages are represented in training and for a deployment whose entire lineage is held out.

All results come from a single evaluation run in which the untrained and trained policies were scored on the same scenarios. We report the deployment groups separately rather than averaging them because they differ in lineage relationship, procedure complexity, and scenario count.

\subsection{Experimental Setup}

The evaluation suite contains 43 scenarios across four deployments, with 21, 8, 8, and 6 scenarios respectively. Each scenario contains a fixed customer script and an expectation block. Five of the 43 scenarios also specify tools that must not be called in that case. None of these negative tool expectations is violated by either arm reported here (Section~\ref{sec:test}).

All four deployment procedures are held out from training, and the evaluation shard contains no real company data (Section~\ref{sec:population}). The deployments differ in their relationship to the training population: one deployment comes from the completely withheld lineage, while the other three have related lineages represented in the training shard.

The four procedures also differ in complexity. The largest declares 15 states, 7 tools, 35 constraints, and 57 catalogue entries, while the smallest declares 11 states, 2 tools, 8 constraints, and 22 catalogue entries. We therefore report each deployment separately rather than treating every scenario as being drawn from the same evaluation distribution.

\subsubsection{Classification of checks}
\label{sec:checkclasses}

A \textbf{documented check} evaluates behaviour whose expected condition is determined by the procedure. In this evaluation, these checks include whether the required tools are called and whether the terminal disposition matches the procedure.

A \textbf{grounded check} evaluates behaviour whose expected condition is determined by the observed conversation. These checks include whether tool arguments match information provided by the caller and whether emitted beats are allowed by the procedure path taken by that conversation.

The classification is operational rather than ontological: a check is assigned according to which source determines the reference condition used by the scorer. Many behaviours depend on both the procedure and the conversation; the distinction concerns where the expected value comes from. Table~\ref{tab:vii} in Appendix~\ref{app:checks} lists every check, its classification, and its denominator.

The two grounded checks, \texttt{args} and \texttt{no\_off\_flow\_beats}, are defined with respect to the structured action space in Section~\ref{sec:env}. Because the policy selects identifiers from a per-example shuffled catalogue, \texttt{no\_off\_flow\_beats} checks whether each selected entry is permitted by some state on the conversation's procedure path, together with entries that the procedure allows independently of a particular state. The check is performed over the resulting set of emitted beats rather than turn by turn. Beat ordering is evaluated separately.

This distinction is important because the policy's spoken responses are catalogue selections rather than generated text. The evaluation therefore does not judge linguistic phrasing.

\subsubsection{Procedure-derived supervision and evaluation}

The documented checks use each deployment's own procedure, including its declared tools and terminal dispositions. The held-out deployment's procedure is not present in training. However, both the trainer and the scorer consume the same procedure representation through the common loading interface described in Section~\ref{sec:representation}.

The results should therefore be interpreted as a decomposition of measured behaviour rather than evidence that the model has learned an abstract concept of ``procedure.'' The shared specification also introduces a potential evaluation blind spot: a defect shared by the supervision and evaluation process may not be detected by the resulting metric. We examine this issue in Section~\ref{sec:discussion}.

\subsubsection{Scoring policy actions}

Every action emitted by the policy is scored, including actions that would not have taken effect in the environment. The evaluation therefore measures the policy's emitted behaviour rather than only the actions that successfully executed downstream.

\subsubsection{Decoding and repeatability}
\label{sec:repeatability}

Policy sampling is disabled using temperature 0. The runtime is nevertheless not fully deterministic. Repeated runs with the same prompt and weights produced 8/21 and 9/21 on one suite. This difference reflects run-to-run measurement variation and is not treated as statistical uncertainty.

The comparisons below are therefore reported as observed outcomes from the evaluation run rather than as claims about the variability of the underlying policy. Procedure generation, scenario worlds and the simulation date are seeded (Appendix~\ref{app:training}); the serving runtime is not.

\subsubsection{Common evaluation session}

The evaluation arms were served and scored in the same session using the same evaluation suites, procedure specifications, and derived evaluation keys. No result was carried over from an earlier evaluation run.

\subsection{Two Behavioural Components Have Different Starting Points}

\begin{figure*}[t]
  \centering
  \includegraphics[width=0.82\textwidth]{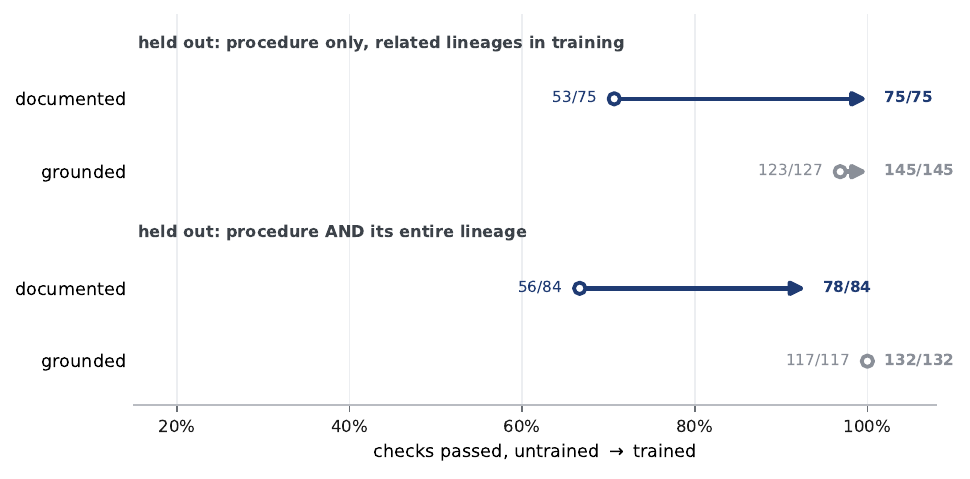}
  \caption{Each arrow runs from the untrained to the trained policy. The documented rows start lower and travel further; the grounded rows start at or near ceiling.}
  \label{fig:twohalves}
\end{figure*}
Table~\ref{tab:iii} separates documented and grounded checks, while Fig.~\ref{fig:twohalves} shows the change from the untrained to the trained policy.

Before training, the 9B policy already passes 97\% and 100\% of grounded checks in the related-lineage and whole-lineage groups, respectively. Its performance on documented checks is lower: 71\% when related lineages are represented in training and 67\% when the entire lineage is held out.

After training, documented performance increases to 100\% on the related-lineage deployments and 93\% on the whole-lineage holdout. Grounded performance reaches 100\% in both groups.

\begin{table}[h]
\centering
\caption{Checks passed, grouped by the source that determines the reference condition. The two documented checks and two grounded checks are reported here. Declared constraints and required-beat coverage are reported separately in Table~\ref{tab:iv}. Grounded denominators differ between arms because they count actions the policy emitted. Checks within a scenario are not independent.}
\label{tab:iii}
\footnotesize
\begin{tabular}{@{}llrr@{}}
\toprule
Holdout & Check type & Untrained & Trained \\
\midrule
Related lineage & Documented & 53/75 (71\%) & 75/75 (100\%) \\
Related lineage & Grounded & 123/127 (97\%) & 145/145 (100\%) \\
Whole lineage & Documented & 56/84 (67\%) & 78/84 (93\%) \\
Whole lineage & Grounded & 117/117 (100\%) & 132/132 (100\%) \\
\bottomrule
\end{tabular}
\end{table}

The main difference between the two groups is their starting point. The untrained policy is already strong on grounded checks, leaving relatively little room for improvement. The largest measured improvement is therefore on documented checks, whose reference conditions are explicitly determined by the procedure.

The customer side of the evaluation is simulated. Customer event paths are sampled from the procedure's own event graph, so the grounded results measure behaviour under the generated evaluation distribution rather than general conversational competence. In particular, these results do not establish that the policy matches the distribution of production customer conversations.

\subsection{What Training Changes}

Table~\ref{tab:iv} reports the individual evaluation dimensions.

\begin{table}[h]
\centering
\caption{Evaluation dimensions for the two holdout classes.}
\label{tab:iv}
\footnotesize
\begin{tabular}{@{}llrr@{}}
\toprule
Holdout & Dimension & Untrained & Trained \\
\midrule
Related lineage & Tools required by state & 42/53 & 53/53 \\
Related lineage & Disposition & 11/22 & 22/22 \\
Related lineage & Declared constraints & 95/96 & 96/96 \\
Related lineage & Required beats covered & 79/86 & 86/86 \\
Related lineage & Staying on declared path & 79/83 & 86/86 \\
Related lineage & Arguments grounded & 44/44 & 59/59 \\
Related lineage & Scenarios passed & 7/22 & 22/22 \\
Whole lineage & Tools required by state & 45/63 & 60/63 \\
Whole lineage & Disposition & 11/21 & 18/21 \\
Whole lineage & Declared constraints & 125/126 & 126/126 \\
Whole lineage & Required beats covered & 67/71 & 71/71 \\
Whole lineage & Staying on declared path & 71/71 & 71/71 \\
Whole lineage & Arguments grounded & 46/46 & 61/61 \\
Whole lineage & Scenarios passed & 8/21 & 18/21 \\
\bottomrule
\end{tabular}
\end{table}

The documented dimensions show substantial improvement in both holdout groups. Required-tool coverage increases from 42/53 to 53/53 on related-lineage deployments and from 45/63 to 60/63 on the whole-lineage holdout. Disposition accuracy increases from 11/22 to 22/22 and from 11/21 to 18/21, respectively.

Declared constraints do not separate the two arms in this evaluation: performance is already close to the ceiling before training, increasing from 95/96 to 96/96 and from 125/126 to 126/126.

The grounded dimensions also improve, but they begin near ceiling. This limits the amount of measurable improvement available. The result therefore indicates that the training run had little room to improve these checks rather than establishing that grounded behaviour is unaffected by training.

Argument accuracy requires additional care in interpretation because its denominator depends on the number of tool calls emitted by the policy. The underlying argument check combines two conditions: conformance to the procedure's declared argument schema and agreement with information supplied by the caller. Only the latter is grounded in the sense used in this paper. In this evaluation, the failing untrained argument observations are format violations, which are counted within the combined argument dimension.

Required-beat evaluation can also be decomposed into coverage and ordering. In every cell the ordered count equals the coverage count --- 79/86 untrained and 86/86 trained for related lineages, 67/71 and 71/71 for the whole-lineage holdout --- so no required beat was ever produced out of order. The observed change in the beat dimension therefore comes from coverage rather than ordering.

\subsection{Per-Deployment Results}

Table~\ref{tab:v} reports scenario-level pass rates for each deployment. A scenario passes only when every check associated with that scenario passes.

\begin{table}[h]
\centering
\caption{Scenario-level pass counts by deployment.}
\label{tab:v}
\footnotesize
\begin{tabular}{@{}lrrr@{}}
\toprule
Deployment & Scenarios & Untrained & Trained \\
\midrule
Related-lineage A (debt) & 8 & 0 & 8 \\
Related-lineage B (debt) & 8 & 3 & 8 \\
Related-lineage C (non-debt) & 6 & 4 & 6 \\
Whole-lineage holdout (debt) & 21 & 8 & 18 \\
\midrule
\textbf{Total} & \textbf{43} & \textbf{15} & \textbf{40} \\
\bottomrule
\end{tabular}
\end{table}

The scenario-level results are not uniform across deployments. Across the three related-lineage deployments, 7 of 22 scenarios pass before training and all 22 pass after training. The whole-lineage holdout increases from 8/21 to 18/21. The non-debt deployment changes from 4/6 to 6/6 and therefore has less room for improvement.

The non-debt deployment differs from the other deployments in several respects: it is the only non-debt call type, has the smallest suite, and its closest related training lineage belongs to a different call type. The current experiment does not isolate which of these factors explains its higher untrained starting point. We therefore do not generalize this per-deployment pattern beyond the four evaluated deployments.

\subsubsection{Tool-surface variation}

Tool-name variation is also not uniform across the four deployments. Two deployments use canonical debt-tool names that also occur in the training shard. The whole-lineage holdout does not preserve the same role-based tool representation because its specification does not record a tool's role in the same form (Appendix~\ref{app:training}). This difference should be considered when interpreting the tool-coverage results.

\subsubsection{Paired scenario comparison}

Each scenario is evaluated by both arms, allowing paired comparison at the scenario level. On the related-lineage block, 15 of 22 scenarios improve and none regress, giving an exact McNemar test of \(p=6.1\times10^{-5}\). On the whole-lineage holdout, 10 of 21 improve and none regress, with \(p=2.0\times10^{-3}\). Across all 43 scenarios, 25 improve and none regress, with \(p=6.0\times10^{-8}\).

These tests apply to paired scenario outcomes. Checks within a scenario are not independent, so the per-dimension counts remain the primary evidence for the behaviour-level analysis.

\subsection{Effect of Model Scale}

The client hardware described in Section~\ref{sec:introduction} provides a single 40\,GB accelerator, which holds a 9B model at 16-bit precision but not a 27B one. We therefore focus the main experiment on the 9B policy.

For context, we also evaluate one untrained 27B checkpoint under the same evaluation specification and key. The 9B policy passes 15/43 scenarios before training and 40/43 after training, while the untrained 27B checkpoint passes 37/43 (Table~\ref{tab:vi}).

\begin{table}[h]
\centering
\caption{Scenario-level pass counts under the same evaluation specification and key.}
\label{tab:vi}
\footnotesize
\begin{tabular}{@{}lr@{}}
\toprule
Arm & Scenarios passed \\
\midrule
Untrained 9B & 15/43 \\
Trained 9B & 40/43 \\
Untrained 27B & 37/43 \\
\bottomrule
\end{tabular}
\end{table}

This 27B result is a single additional checkpoint rather than a second controlled training arm. The 43-scenario suite is also relatively small for drawing conclusions about model scale. We therefore use it only as contextual evidence that an untrained larger model can perform substantially better than the untrained 9B on this suite. The experiment does not establish how training effects vary with model size.

Overall, the main experiment supports a more specific conclusion: within the evaluated 9B policy, training across the generated procedure population substantially increases performance on documented checks and scenario-level outcomes, including on the single held-out lineage. Grounded checks begin near ceiling in this evaluation, leaving substantially less room for measurable improvement.

\section{Discussion}
\label{sec:discussion}

\subsection{What This Means for a Policy Trained Once and Shipped}

The main question of this study is whether a single small policy can be trained once and then follow procedures that were not seen during training. The results provide evidence for this setting. On the completely held-out lineage, documented behaviour improves after training: required-tool coverage increases from 45/63 to 60/63, while disposition accuracy increases from 11/21 to 18/21. The same pattern is observed for the three deployments whose related lineages are represented in training, where documented checks reach 100\% after training.

In contrast, grounded behaviour is already strong before training, reaching 97--100\% across the two deployment groups. Training therefore produces its largest measured improvement on behaviour whose expected condition is explicitly determined by the procedure. This supports the decomposition defined in Section~\ref{sec:checkclasses}: procedure following is not a single capability, and the main training gain is concentrated in the documented component.

The untrained 27B checkpoint provides an important qualification. It passes 37/43 evaluation scenarios without procedure-derived training, compared with 15/43 for the untrained 9B. Thus, procedure-following capability is not absent from a larger pretrained model. The contribution of the present training setup should instead be understood as improving the smaller 9B policy within the deployment constraints of the study. Because no controlled ablation was performed, the results do not isolate which component of the training design produces this improvement.

The transfer result should also be scoped carefully. Only one complete lineage is withheld from training, and the evaluation contains four deployment procedures. The study therefore demonstrates transfer to the tested held-out lineage rather than establishing general transfer across arbitrary procedures. Similarly, the experiment evaluates unseen procedures but does not test whether a trained policy can adapt to a revised version of a procedure it has already seen.

\subsection{What the Evaluation Measures}

The reported results should be interpreted as conformance to the formalised procedure rather than correctness against an independently constructed ground truth. The same specification is used to generate the supervision, constrain the environment, and derive the evaluation criteria. This shared dependency is central to the design, but it also creates a potential blind spot: a defect in the specification can propagate into both training and evaluation without necessarily appearing as a model error.

For this reason, the evaluation should be understood as measuring how consistently the policy follows the behaviour formalised by the specification. This distinction is particularly important for regulated applications, where the correctness of the procedure itself is part of the system being evaluated. We therefore separately audit the specification and its derived evaluation artefacts. The audit identified 11 specification-related defects, including 3 that were not exposed by the corresponding evaluation metric (Appendix~\ref{app:audit}). One exposed defect shows what is at stake: in about 1\% of generated episodes, a repair step disclosed the balance to a caller whose identity had not been verified, which is the disclosure that the Thai Debt Collection Act prohibits \cite{thaidca2015}. A defect in the specification pipeline can therefore put legal violations into the training data. These findings do not change the results in Section~\ref{sec:experiments}; they qualify what those results can establish.

The implication is that specification-conditioned training requires validation at two levels: whether the policy conforms to the specification, and whether the specification accurately represents the intended behaviour. The latter cannot be established solely by measuring policy performance against tests derived from the same specification.

\subsection{Limitations}

Several limitations constrain the conclusions of this study.

First, conformance to \flowspec{} is not legal compliance. The specification encodes one creditor's reading of the law, and only 8 of the 35 constraints in the representative deployment are typed and checked (Section~\ref{sec:representation}). Rules that govern when a call is placed, such as permitted hours and call frequency, are outside the policy's scope. Evaluation is also on text transcripts with a simulated caller, so speech-recognition errors and real callers' behaviour are not modelled.

Second, the study does not isolate the contribution of individual components of the training design. We do not have ablations for procedure conditioning, generated trajectories, per-example randomization, training-set size, or additional supervised fine-tuning. We also do not compare directly with a separately trained policy for each procedure or with a runtime procedure-prompting approach under the same deployment constraints.

Third, procedure revision is not evaluated. The experiment tests transfer to procedures that are unseen during training, including a completely withheld lineage, but it does not test adaptation to modifications of procedures that were previously seen.

Fourth, the evaluation covers four deployments and one completely withheld lineage. The deployment procedures also differ in their tool surfaces and other structural properties, so the experiment cannot isolate lineage transfer from all other sources of variation.

Fifth, the evaluation has limited case-level coverage and some runtime variability. The scenario-level denominators are small, and repeated runs with the same prompt and weights produced different outcomes in one suite. The reported values should therefore be read as observed results from the evaluation run rather than as precise estimates of deployment-wide performance. In addition, the evaluation-session provenance does not fully resolve the model path used for the original run, so checkpoint identity relies partly on run records and subsequent audit information.

Finally, the study focuses on a 9B policy trained with supervised fine-tuning and does not provide a controlled comparison across model scales or training methods. The 27B result is an additional untrained contextual reference rather than a matched training arm. Reinforcement learning, quantization effects, and production compliance controls are outside the scope of the present experiment.

Overall, the results support a narrower conclusion than general-purpose procedure following: a single small policy can be trained across a population of machine-readable procedures and can improve its documented behaviour on procedures from an unseen lineage, while already-strong grounded behaviour remains near ceiling. At the same time, the measured performance is bounded by the quality of the specification and evaluation artefacts used to define that behaviour.

\section{Conclusion}
\label{sec:conclusion}

We presented a specification-conditioned dialogue policy for voice agents whose conversational flow is fixed by law, using outbound debt collection as the case study: the agent must verify identity before disclosure, make mandated disclosures, respect payment-request limits and record a correct disposition, while conversations cannot be sent to a hosted model and spoken responses must come from an approved catalogue. Rather than treating the procedure as external context, we place it inside the training and evaluation loop. A single machine-readable procedure per deployment generates the policy instruction, constrains the environment and its tools, defines the reference behaviour used for supervision, and specifies the corresponding evaluation checks. Training across a population of such procedures, with complete lineages held out, allows us to evaluate transfer to unseen procedures.

The results show a clear difference between two components of measured procedure following. Documented checks, whose expected conditions are determined by the procedure, are substantially weaker before training and improve to 93--100\% after training. This improvement is also observed for the deployment whose entire lineage is withheld. Grounded checks, whose expected conditions depend on the observed conversation, are already strong before training and reach 100\% after training. Thus, within the scope of this experiment, the main measured gain from training is concentrated in behaviour explicitly determined by the procedure.

These findings support the feasibility of training a single small policy across multiple procedures, but they do not establish that every component of the proposed design is necessary or that the result generalizes to arbitrary procedures. The study does not isolate the contribution of individual training components, does not evaluate procedure revisions, and covers a limited number of deployments and withheld lineages.

Finally, the evaluation itself must be interpreted with care. Because the same specification provides the supervision and evaluation criteria, the reported scores measure conformance to the formalised procedure rather than correctness against an independent ground truth. Specification defects can therefore remain invisible to the metric derived from that specification. This highlights an important requirement for specification-conditioned agents: evaluating the policy is not sufficient; the specification and its derived evaluation artefacts must also be validated.

Overall, the study demonstrates a concrete form of transfer: a single 9B policy trained across machine-readable procedures can improve its documented procedure-following behaviour on procedures from an unseen lineage, while maintaining strong grounded behaviour. The results also define the boundary of that claim and motivate future work on component ablations, procedure revision, broader lineage coverage, and independently grounded evaluation.

\bibliographystyle{IEEEtran}
\bibliography{refs}

\appendices
\section{Procedure Schema and Action Space}
\label{app:spec}

\textbf{Schema.} Six top-level keys are required: \texttt{flow\_id}, \texttt{events}, \texttt{tools}, \texttt{states}, \texttt{faq\_routing} and \texttt{constraints}. The schema recognizes eleven constraint types, including \texttt{max\_occurrences}, \texttt{once\_per\_call}, \texttt{forbid\_after\_event}, \texttt{require\_tool\_before\_end} and \texttt{tool\_pair}; each constraint declares its enforcing layers as a subset of \{prompt, reward, backend, session\}.

\textbf{Validation.} The validator rejects a procedure, rather than warning, on: a transition to a nonexistent state or on an undeclared event; duplicate transitions for one event; a non-terminal state with no transitions; an unknown constraint type or invalid enforcement layers; tool gating that references an undeclared tool or state; and a \texttt{fine\_state} that resolves to no catalogue entry. Unreachable states and unbound catalogue entries are warnings.

\textbf{Example.} A gated tool declaration, abbreviated:
{\footnotesize
\begin{verbatim}
name: check_account_status
gating:
  after_event: name_confirmed
  required_before_state: disclose_ask
  max_calls_per_conversation: 1
\end{verbatim}
}
The tool cannot be called before the caller confirms their name, must be called before the state that discloses the balance, and may be called once per call. This is how the legal rule ``verify identity before disclosure'' becomes machine-checkable.

\textbf{Action space.} Every declared tool is presented to the policy as a function schema built from its declaration. One further tool is appended for speech:
{\footnotesize
\begin{verbatim}
name: reply
parameters:
  text_ids:
    type: array
    items:
      enum: [ids]
\end{verbatim}
}
The enumeration holds exactly the catalogue identifiers of that conversation, after per-example renumbering and shuffling. The environment maps each identifier to its approved sentence and fills call-time slots from its own records; the policy supplies no part of the sentence. The training builder and the serving task generator construct this schema with the same code.

\section{The Evaluation Checks}
\label{app:checks}

Table~\ref{tab:vii} lists the twelve checks, the class each is assigned according to the source of its reference condition, and its denominator. The documented and grounded checks are aggregated in Table~\ref{tab:iii}; the two dimension checks are reported in Table~\ref{tab:iv}; validity checks verify the generated evaluation artefacts and are not aggregated.

\begin{table*}[!ht]
\centering
\caption{The twelve evaluation checks, their class, and what each is counted against.}
\label{tab:vii}
\footnotesize
\begin{tabular}{@{}lll@{}}
\toprule
Check & Class & Denominator / Reference \\
\midrule
\texttt{tools} & documented & Tools required by the state \\
\texttt{outcome} & documented & Scenarios that must record an outcome \\
\texttt{no\_off\_flow\_beats} & grounded & Beats emitted during the conversation \\
\texttt{args} & grounded & Tool calls emitted by the policy \\
\texttt{beats} & dimension & Required beats, evaluated per scenario \\
\texttt{spec\_constraints} & dimension & Typed constraints evaluated \\
\texttt{forbidden} & diagnostic / case-level & Scenarios naming a forbidden tool \\
\texttt{no\_empty\_slots} & validity & Render validity; must be zero \\
\texttt{short\_script} & validity & Script validity; must be zero \\
\texttt{suite\_vs\_spec} & validity & Suite/spec consistency; must be zero \\
\texttt{spec\_walk} & validity & Procedure-walk validity; must be zero \\
\texttt{must\_not\_declared} & validity / diagnostic & Consistency of negative declarations \\
\bottomrule
\end{tabular}
\end{table*}

\begin{table*}[!ht]
\centering
\caption{Specification audit findings.}
\label{tab:viii}
\footnotesize
\begin{tabularx}{\textwidth}{@{}rXlc@{}}
\toprule
\# & Defect & Scope & Exposed by metric \\
\midrule
1 & Generated episodes never recorded the customer's agreement event & One generation pattern & yes \\
2 & A beat without a template inherited the wording of another beat & One intent & yes \\
3 & Three independent faults in one deployment: constraints without identifiers caused generation to fail, one declared tool was not implemented, and another declared no argument map & Whole deployment & yes \\
4 & A deployment declared no disposition tool, but the metric required one & Whole deployment & yes \\
5 & Alternative templates were collapsed into a single response containing all alternatives & One state & yes \\
6 & A repair example injected an amount that the caller had never stated & 15\% of episodes & yes \\
7 & A balance was disclosed to an unverified caller, a data-protection violation & $\sim$1\% of episodes & yes \\
8 & 56 identifiers were hard-coded in prose even though identifiers are shuffled per example & 8 procedures and 2 evaluation specifications & yes \\
9 & Silence was routed to a success disposition & 3 procedures & no \\
10 & Data fields were used as scratch space & 3 procedures & no \\
11 & Generator internals were written into a customer-facing field & 441 of 970 recorded calls & no \\
\midrule
 & \textbf{Total} & \textbf{11 defects} & \textbf{8 exposed $\cdot$ 3 not exposed} \\
\bottomrule
\end{tabularx}
\end{table*}

The \texttt{args} denominator depends on how many tool calls the policy emits, so it is reported as a count rather than a rate. The \texttt{forbidden} check is the only case-level check declared by the scenario rather than the procedure; it passes in all 43 scenarios for both arms.

\section{Training Configuration and Seeding}
\label{app:training}

The policy is Qwen3.5-9B with a rank-32 LoRA adapter ($\alpha = 64$, dropout 0) on all linear layers. Maximum sequence length is 16,384 tokens, and over-length examples raise an error rather than being truncated. Micro-batch size is 1 and effective batch size 16. The learning rate is $2\times10^{-4}$ with a cosine schedule and 3\% warmup. Training runs for three epochs (468 steps) in bfloat16 on one H100 for about five hours; final training loss is 0.022 and validation loss 0.068.

The population of 340 procedures is generated with seed 42; tool-name pools are keyed by each tool's declared role, which one lineage's specification does not record, so that lineage keeps its base tool names. Scenario worlds (balance, due date, identity digits) are seeded by a 32-bit cyclic redundancy check (CRC32) hash of the scenario identifier, and all scoring uses a fixed simulation date. Decoding uses temperature 0, but the runtime is not fully deterministic (Section~\ref{sec:repeatability}).

\section{Specification Audit}
\label{app:audit}

Table~\ref{tab:viii} lists each defect found in the procedure definitions and their derived evaluation artefacts, its scope, and whether the derived metric exposed it. The three unexposed defects were shared by the specification, the supervision and the test, so the policy's behaviour appeared compliant. Defect 7, a balance disclosed to an unverified caller, is the kind of disclosure that the Thai Debt Collection Act prohibits (Section~\ref{sec:introduction}). The audit qualifies the scores in Section~\ref{sec:experiments} and does not change them.

\end{document}